\documentclass[runningheads]{llncs}
\usepackage{graphicx}
\usepackage{amsmath, amsfonts, amssymb}
\usepackage{xspace}
\usepackage{stmaryrd}
\usepackage{hyperref}
\usepackage[T1]{fontenc}
\usepackage[utf8]{inputenc}
\usepackage{graphicx}
\usepackage{listings}
\usepackage{subcaption}
\usepackage{xcolor}
\usepackage{float}
\usepackage{lstautogobble}
\usepackage{color}
\usepackage{proof}
\usepackage{multicol}
\usepackage{mathpartir}
\usepackage{todonotes}
\usepackage{cleveref}
\usepackage{mathpartir}
\usepackage{wrapfig}
\usepackage{mathtools}
\usepackage{tabularray}
\UseTblrLibrary{booktabs}
\usepackage{orcidlink}
\usepackage{marvosym}
\usepackage{array}

\newboolean{fullversion}
\setboolean{fullversion}{true}

\usetikzlibrary{fit,arrows.meta,calc}

\tikzset{
  ->/.style={-{Stealth}, shorten >=0.1mm},
  <-/.style={{Stealth}-, shorten <=0.1mm},
}

\newcommand{\prob}{\mathbb{P}}

\definecolor{bluekeywords}{rgb}{0.13, 0.13, 1}
\definecolor{greentypes}{rgb}{0, 0.5, 0}
\definecolor{inferedgreentypes}{rgb}{1.0, 0.2, 0}
\definecolor{orangecomments}{rgb}{1, 0.5, 0.1}
\definecolor{redstrings}{RGB}{171, 114, 2}
\definecolor{graynumbers}{rgb}{0.5, 0.5, 0.5}
\definecolor{goldcomments}{rgb}{0.6, 0.4, 0.08}

\lstdefinelanguage{Lola}{
  keywords=[0]{input, output, trigger, constant, import, spawn, eval, close, with, when},
  moredelim=**[is][\transparent{0.6}]{?}{?},
  moredelim=**[is][\color{greentypes}@]{@}{@},
  keywordstyle=[0]\bfseries\color{bluekeywords},
  keywords=[1]{if, then, else, aggregate, defaults, offset, last, by, or, to, sin, cos, abs, hold, over, using, over_instances, prob, prior, confidence, given},
  keywords=[2]{Variable, String, Int, Int64, UInt, UInt64, Bool, Float32, Float64, Float, Time},
  keywordstyle=[2]\color{greentypes},
  sensitive=false,
  comment=[l]{//},
  morecomment=[s]{/*}{*/},
  morestring=[b]',
  morestring=[b]",
  literate={\\@}{@}1
}
\makeatletter
\begin{document}
\title{FairMon: A Tool for Monitoring and\\Visualizing Algorithmic Fairness}
%
%
\author{
 Jan Baumeister\inst{1}\orcidlink{0000-0002-8891-7483} \and
 Bernd Finkbeiner\inst{1,2}\orcidlink{0000-0002-4280-8441} \and
 Vladimir Krsmanovic\inst{1}\orcidlink{0009-0004-5892-0421} \and\\
 Frederik Scheerer\textsuperscript{(\Letter)}\inst{1}\orcidlink{0009-0007-8115-0359} \and
 Julian Siber\inst{1}\orcidlink{0000-0003-0842-0029} \and
 Tobias Wagenpfeil\inst{1}\orcidlink{0009-0005-2680-1150}}
 \authorrunning{J.\ Baumeister et al.}
%
 \institute{CISPA Helmholtz Center for Information Security, Saarbrücken, Germany\newline
 \email{\{jan.baumeister,finkbeiner,vladimir.krsmanovic,\\frederik.scheerer,julian.siber,tobias.wagenpfeil\}@cispa.de}
 \and
 Technical University of Munich, Germany
 }
%
\maketitle              

\begin{abstract}
Runtime monitoring has recently been proposed as a rigorous method for analyzing algorithmic fairness of autonomous decision systems used in critical scenarios such as credit lending, job application, and the criminal justice system.
Prior work has shown that runtime monitoring, in principle, can be an effective technique for establishing the kind of human oversight required by legislation such as the EU Artificial Intelligence Act.
In practice, the available monitoring tools have not been developed with this application in mind and display several critical shortcomings in these scenarios.
In this paper, we present FairMon, a runtime monitoring tool tailored to fairness analysis of high-stakes decision systems.
FairMon uses RTLola as a flexible specification language for monitors, which we have extended with conditional probability operators that allow for concise descriptions of algorithmic fairness properties.
The tool also features a real-time visualization of intermediary values, enabling human insight into the dynamics of the monitored system. 

\end{abstract}

\section{Introduction}

Algorithmic fairness requires that systems act without bias against protected social groups~\cite{DworkHPRZ12}.
Several inquiries have shown that critical autonomous decision systems often fail to adhere to this requirement.
A prominent example is the COMPAS tool for recidivism risk prediction, which has displayed highly diverging true and false positive rates between Black and White Americans when predicting a high risk of recidivism~\cite{ProPublica16}.
Hence, the system violated the fairness requirement known as \emph{equalized odds}~\cite{HardtPNS16}, which requires that the true and false positive rates for an outcome $Y$ are within some bound $\epsilon$ between all groups:
$$\big| \, \prob (\hat{Y} = 1 \mid G = 1, Y = 1) - \prob (\hat{Y} = 1 \mid G = 0, Y = 1) \, \big| \leq \epsilon \enspace .$$
In this true positive part of the property, $\hat{Y}$ denotes the system's prediction, corresponding to a high-risk assessment in the COMPAS setting, and $Y$ is the true outcome, indicating whether an individual actually reoffended.
$G$ encodes membership in a protected group.

Runtime monitoring is a well-established technique in the formal methods community for analyzing executions of diverse high-stakes systems such as unmanned aircraft~\cite{DBLP:conf/cav/BaumeisterFKLMST24,DBLP:conf/cav/JohannsenJKRZ23,DBLP:conf/fm/PerezGD24} and distributed networks~\cite{FaymonvilleFST16}.
Several recent studies have also highlighted its general applicability for monitoring algorithmic fairness~\cite{HenzingerKKM23a,HenzingerKM23}.
In a recent case study~\cite{DBLP:conf/tacas/BaumeisterFSSW25}, we have shown that stream-based monitoring with the specification language RTLola~\cite{BaumeisterFKS24} can monitor algorithmic fairness requirements in decision and prediction systems over time.
Building on this exploratory case study, we have developed a dedicated tool for this purpose: FairMon.
It addresses shortcomings of existing stream-based monitoring frameworks, including RTLola, with respect to syntactical succinctness and the visualization during execution.

\paragraph{Dedicated Operators.}
Encoding algorithmic fairness with RTLola as described in~\cite{DBLP:conf/tacas/BaumeisterFSSW25} requires elaborate usage of parameterized streams and aggregation.
Not only is this an error-prone process even for experts, but it also results in a high threshold for adoption by practitioners.
To make the specification of fairness properties more intuitive, we have added probability operators to RTLola.
These directly encode (conditional) probabilities of events and, in this way, allow a direct encoding of properties such as equalized odds.
Assuming the boolean variables are inputs, specifying this requirement with the original approach~\cite{DBLP:conf/tacas/BaumeisterFSSW25} required a minimum of five output streams.
With the dedicated operators, all that is needed is the following specification, which mirrors the mathematical formulation of the equalized odds property:
\begin{lstlisting}
    trigger abs(y_hat.prob(given: g$\land$$\neg$y) - y_hat.prob(given: $\neg$g$\land$$\neg$y)) > $\epsilon$
\end{lstlisting}
\begin{figure}[t]
    \centering
    \includegraphics[width=\linewidth]{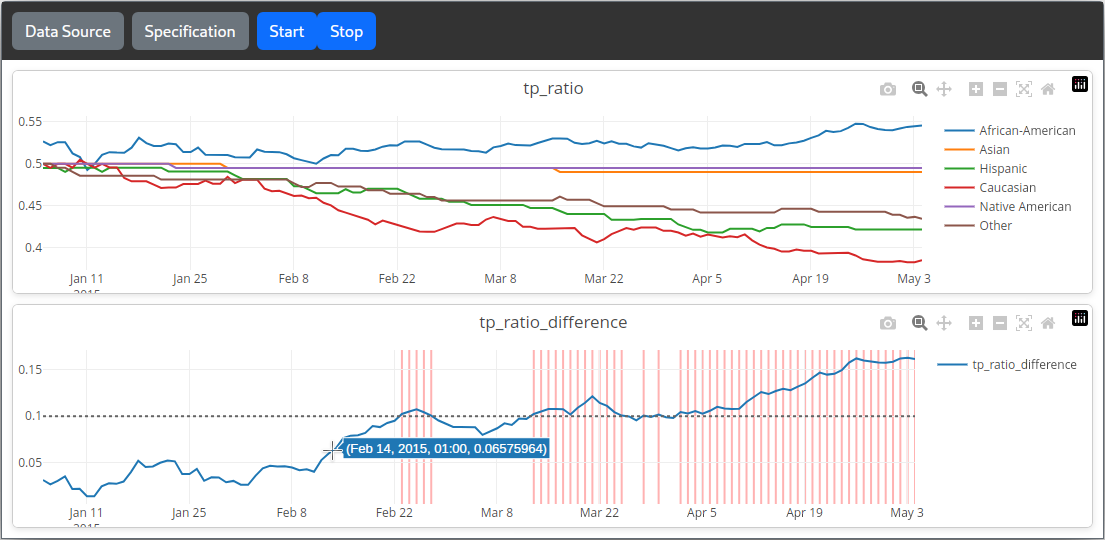}
    \caption{Screenshot of FairMon monitoring the COMPAS dataset.}
    \label{fig:tool_screenshot}
\end{figure}
This condition is evaluated at runtime by the monitor, and a warning is issued as soon as the trigger condition evaluates to true.
Accordingly, the trigger condition is defined as the negation of the equalized odds property above.
\paragraph{Visualization.}
In addition, FairMon comes with a graphical user interface, depicted in \Cref{fig:tool_screenshot}, that makes the monitoring process more accessible.
The GUI enables the configuration of data sources, the editing of fairness specifications, and the visualization of the monitoring process through line plots of the important values.
This allows the system's state and fairness to be tracked over time, with violations easily detectable in a visual manner.

\paragraph{Integration.}
FairMon is designed to integrate easily into existing decision-making systems through an integration framework.
Its interface allows users to choose and configure input sources, supporting both domain-specific and general-purpose data sources.
To demonstrate this flexible interface, we connect FairMon to the EU's DSA Transparency Database API~\cite{ec_dsa_transparency_2023}, which provides a stream of moderation decisions of large online platforms and offers an example for a large-scale decision-making system.


\paragraph{Related Work.}
There exists a broad range of literature on the theory of algorithmic fairness; we refer to \cite{PessachS23,BarocasHN23} for a comprehensive overview.
Runtime monitoring of algorithmic fairness has been studied in recent work~\cite{HenzingerKKM23a,HenzingerKKM23b,HenzingerKM23,DBLP:conf/tacas/BaumeisterFSSW25}.
While some approaches exist that present specific syntax for expressing fairness properties~\cite{zhang2021omnifair,morales2024langbite}, these primarily target machine learning or test generation rather than runtime monitoring.
Various methods exist for visualizing the execution of runtime monitors, whether for offline analysis~\cite{DBLP:conf/atva/FinkbeinerKS23,aurandt2025r2u2,DBLP:conf/rv/KallwiesLSSTW22} or online scenarios~\cite{DBLP:conf/rv/BaumeisterFGS22,baumeister2024rtlolamo3vis}.
While visualization tools for fairness exist in the area of machine learning~\cite{DBLP:conf/ieeevast/CabreraEHKMC19,DBLP:journals/corr/abs-1811-05577}, they do not offer the flexibility of stream-based specification languages.

\section{Probabilistic RTLola}

RTLola~\cite{BaumeisterFKS24} is a stream-based specification language for runtime monitoring.
A specification consists of stream definitions, each stream representing an infinite sequence of values.
\emph{Input streams} capture values reflecting the current state of the monitored system, while \emph{output streams} compute new values by filtering or aggregating other streams.
Violations of the specification are detected through \emph{triggers}, which are special, boolean-valued, output streams: whenever a trigger evaluates to true, a violation of the specification has occurred.

\begin{example}\label{ex:sliding_prob}
Consider the following specification computing the probability of a boolean event over a timed window:
\begin{lstlisting}
  input a : Bool
  output sum_a @10s@ := a.aggregate(over: 60s, using: $\Sigma$)
  output count_a @10s@ := a.aggregate(over: 60s, using: count)
  output prob_a @10s@ := sum_a / count_a
  trigger prob_a > 0.1
\end{lstlisting}
The specification declares a boolean-valued input stream \lstinline!a! and computes two sliding window aggregations over the last 60 seconds.
The first, \lstinline!sum_a!, counts how often \lstinline!a! has been true in that window, while the second, \lstinline!count_a!, counts all events of \lstinline!a!.
Both produce a new value every 10 seconds.
The ratio \lstinline!prob_a! yields the empirical probability that \lstinline!a! is true during the last minute and the trigger raises an alarm if this probability is higher than $0.1$.
\end{example}

\newcolumntype{H}[1]{>{%
  \raggedright
  \hangindent=1em
  \hangafter=1
  \arraybackslash
}m{#1}}

\begin{table}[t]
\centering
\caption{Probability operators in RTLola.}
\label{tab:prob_ops}
\begin{tabular}{@{}H{0.4\linewidth}m{0.58\linewidth}@{}}
\toprule
Syntax & Meaning \\
\midrule
\lstinline!e.prob(given: c)! & $\prob(e \mid c)$ over the entire history \\
\lstinline!e.prob(given: c, over: d)! & $\prob(e \mid c)$ within a window of duration $d$ \\
\lstinline!e.prob(given: c, prior: p, confidence: k)! & as above, smoothed toward prior $p$ with weight $k$ \\
\lstinline!e.prob(given: c, prior: p, confidence: k, over: d)! & as above, windowed and smoothed \\
\lstinline!s.prob(of: $\lambda$)! & fraction of instances of $s$ satisfying $\lambda$ \\
\lstinline!s.prob(of: $\lambda$, given: $\mu$)! & $\prob(\lambda \mid \mu)$ across instances of $s$ \\
\lstinline!s.prob(of: $\lambda$, given: $\mu$, prior: p, confidence: k)! & as above, smoothed \\
\bottomrule
\end{tabular}
\end{table}

Such probability computations are common in fairness specification, as shown in our recent paper~\cite{DBLP:conf/tacas/BaumeisterFSSW25}.
In both that work and the example above, probabilities were represented as floating-point numbers.
This choice, however, limits type checking and makes specifications less precise and harder to interpret.
To address these issues, we introduced a dedicated value type \lstinline!Prob!, representing a probability in the interval $[0,1]$.
Further, the new probability type allows runtime checks to ensure that probability values remain within the valid range of $[0,1]$. It ensures that errors are detected as soon as they occur, rather than letting invalid values propagate through computations. Values outside $[0,1]$ could silently distort fairness measures, leading to incorrect conclusions. 

Building on the new \lstinline!Prob! type, we extended RTLola to include probability-specific operations and aggregations.
These extensions enable the expression of (conditional) probabilities through dedicated operators, allowing for concise formulations of fairness specifications. 

\begin{example}\label{ex:demographic_parity}
As an illustration, consider the definition of demographic parity~\cite{DworkHPRZ12}:
\begin{align*}\big| \, \prob (A = 1 \mid G = 1) - \prob (A = 1 \mid G = 0) \, \big| \leq \epsilon \enspace . 
\end{align*}
With the new operators, this condition translates almost directly into RTLola:
\begin{lstlisting}
    trigger abs(a.prob(given: g) - a.prob(given: $\neg$g)) $>$ $\epsilon$
\end{lstlisting}
\end{example}

In the above example, the shorthand method \lstinline!prob! performs a probability aggregation.
RTLola supports both unconditional and conditional probability aggregations, and allows specifying a prior~\cite{MAPestimation,DBLP:conf/tacas/BaumeisterFSSW25} for when little data is available.
The aggregations can be applied over sliding windows (such as in \Cref{ex:sliding_prob} using the shorthand \lstinline!a.prob(over: 60s)!), over the full history (as in \Cref{ex:demographic_parity}), or across instances of parameterized streams.
A summary of all supported operators is given in \Cref{tab:prob_ops}.

These constructs allow fairness specifications to be expressed far more concisely.
As a result, even non-monitoring experts can formulate fairness specifications naturally.
By aligning the written specification as close as possible to the natural formulation of these fairness notions, we bridge the crucial gap between \emph{what the users want to express} and \emph{how it is expressed}.
Closing this gap is essential for applying fairness toolkits to recommender and decision systems, as well as for monitoring machine learning systems~\cite{DBLP:conf/recsys/BeattieTC22,DBLP:conf/chi/HolsteinVDDW19}.

\section{FairMon}

\begin{figure}[t]
    \centering
    \input{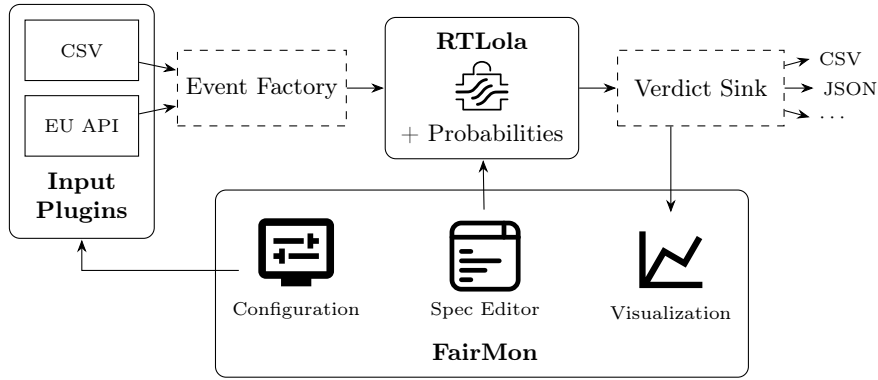}
    \caption{Overview of the FairMon tool.}
    \label{fig:fairmon_overview}
\end{figure}

In addition to concise and understandable specifications, the results produced by the monitor should be easy to interpret and updated continuously at runtime. 
For this purpose, we present FairMon\footnote{Our tool is publicly available at \url{https://github.com/reactive-systems/rtlola-fairmon}.}, a graphical interface for fairness monitoring with RTLola~\cite{BaumeisterFKS24,DBLP:conf/cav/BaumeisterCFS25}.
It provides an intuitive configuration interface for the monitor and enables visualization of its results at runtime.

We show an overview of the FairMon tool in \Cref{fig:fairmon_overview}.
FairMon consists of three components: a specification editor, a configuration interface, and a visualization panel.
To facilitate seamless integration into existing systems, our approach builds on the plugin-based integration framework for RTLola introduced in \cite{DBLP:conf/cav/BaumeisterFKLMST24}.
This framework separates between input and output plugins: input plugins extract input streams from the monitored system through a unified \emph{Event Factory} interface, while output plugins integrate the monitor's output streams into external components using a \emph{Verdict Sink}.
The specification editor allows users to author and modify the RTLola fairness specification, which is passed directly to the monitor. 
Our tool provides a graphical output plugin that visualizes the monitor's internal state and provides live plots of relevant streams, as visible in~\Cref{fig:tool_screenshot}.
Additionally, the graphical interface allows users to configure and select the input plugins for acquiring the data.
Currently, we supply two predefined input plugins, a domain-specific one tailored to the DSA API, and a general-purpose plugin that accepts arbitrary data over CSV. 

\begin{figure}[t]
    \begin{subfigure}[c]{0.48\linewidth}
        \centering
        \includegraphics[width=0.99\linewidth]{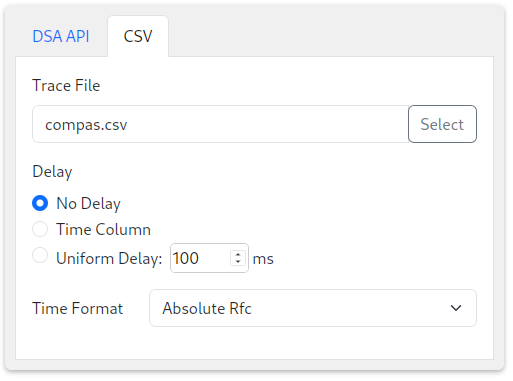}
        \caption{Configuration of the CSV plugin.}
        \label{fig:csv_plugin}
    \end{subfigure}
    \hfill
    \begin{subfigure}[c]{0.48\linewidth}
        \centering
        \includegraphics[width=\linewidth]{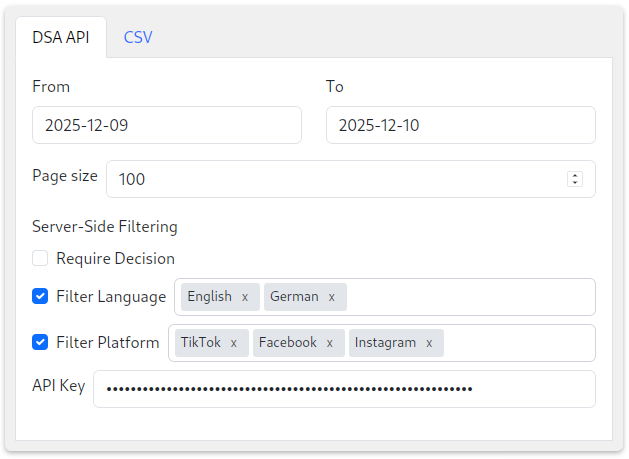}
        \caption{Configuration of the DSA API plugin.}
        \label{fig:api_plugin}
    \end{subfigure}
    \caption{
        Screenshots of the configuration options for different input plugins.
    }
    \label{fig:placeholder}
\end{figure}

To improve the performance of the tool when querying large APIs or datasets, we implement a filtering mechanism to reduce the amount of data loaded. As shown in~\Cref{fig:api_plugin}, the required filters are integrated into our interface, enabling users to configure them directly in the frontend. 
As an example, Instagram emits roughly 5.5 million records per day to the DSA Transparency Database, while often only a fraction of the reported data is of interest.
For example, a specification we present in Section~\ref{sec:examples} requires only a subset of 380,000 records.
Querying the full records is unnecessary and wastes significant time and computing resources due to hitting API limits. 
To combat this type of problem, we expose configuration options for server-side filtering, significantly reducing overall query time.
For the Instagram specification, this mechanism reduces the loading time by 93\%.

For visualizing the stream values over time, we use Plotly, providing users with interactive features such as zooming and panning of the plots.
The configuration of the plots, including which streams are displayed and how they are rendered, is controlled through annotations embedded directly in the RTLola specification.
This annotation mechanism was originally introduced in \cite{DBLP:conf/rv/BaumeisterFS25} and allows for tightly binding the configuration to the specification and preventing the configuration and specification from diverging. 

The visualization tool provides several mechanisms to aid the interpretability of the data and results, all of which can be seen in \Cref{fig:tool_screenshot}.
Triggers are rendered as red vertical bars, offering an immediate visual cue for when conditions are violated, with their associated thresholds represented as horizontal dashed lines.
For parameterized streams, all instances can be displayed within a single plot, while different streams can also be combined in a single plot or displayed across separate plots.
Crucially, all visualization properties are derived directly from the specification itself, requiring no external configuration.
Streams are assigned to plots via an annotated plot identifier, where streams sharing the same identifier are grouped together into a single panel.
\ifthenelse{\boolean{fullversion}}{
Consider \Cref{app:ai_specs} for example specifications.
}{
Consider the full version~\cite{fullversion} of this paper for example specifications.
}


\section{Examples}\label{sec:examples}

We illustrate our tool using two examples: COMPAS and the DSA Transparency API.
The former serves as a well-known example from the literature and allows us to make a direct comparison with specifications from our earlier work.
The DSA Transparency database application uses the newly introduced API from the EU to monitor online moderation data.
\ifthenelse{\boolean{fullversion}}{
All specifications and further details of the experiments can be found in the Appendix.
}{
All specifications and further details of the experiments can be found in the full version~\cite{fullversion} of this paper.
}

\paragraph{COMPAS.}
As a first example, we revisit the equalized odds specification introduced in our previous work~\cite{DBLP:conf/tacas/BaumeisterFSSW25}.
We use the real COMPAS dataset, imported into FairMon via a CSV input plugin as shown in \Cref{fig:csv_plugin}.
The resulting visualization in \Cref{fig:tool_screenshot} displays the true positive rates across groups in the upper plot, and the maximal difference between the groups in the lower plot.
The horizontal dotted line marks the threshold, while the vertical red bars highlight violations of the specification.

The specification must extract multiple independent trials from the inputs, and includes temporal logic to determine whether recidivism occurs within two years of the initial screening.
As a result, the original specification was relatively large, spanning 30 lines.
With the new probability operators, we can express the same behavior far more concisely: the true-positive specification is reduced by 57\% (to 13 lines), and the number of output streams is reduced from 7 to 3.

\paragraph{DSA Transparency API.}
With the introduction of the Digital Services Act (DSA)~\cite{eu_dsa_2022} and the AI Act~\cite{eu_ai_act_2024} by the European Commission, platforms classified as Very Large Online Platforms are legally obligated to report on practices that impact their users. 
One of the obligations under the DSA is that these platforms must provide \emph{statements of reason} following all moderation decisions, which have to contain information about the mechanism used for detection and removal of harmful content, as well as legal reasoning for it.

We extended our tool with an input plugin that interfaces with the DSA database to monitor and visualize relevant fairness properties over these moderation decisions.
In our experiments, we focused on three specifications:
The first examines the ratio of automated decisions across content languages to assess whether groups speaking certain languages face disproportional automated moderation.
The remaining two specifications, adopted from existing work~\cite{trujillo2025dsa}, analyze the delay between content creation and moderation, and highlight the differences between the ratios of automated detection and decision across the different platforms, respectively. We examine the moderation decision made by the Google Play Store during December 2025, with the data being collected from the DSA Statements of Reason database using the API functionality of our tool. 
\ifthenelse{\boolean{fullversion}}{
The specification can be inspected in Appendix~\ref{app:ai_specs}. 
}{
The specification can be inspected in the full version~\cite{fullversion} of this paper.
}

\begin{figure*}[hbt]
    \centering
    \includegraphics[width=\linewidth]{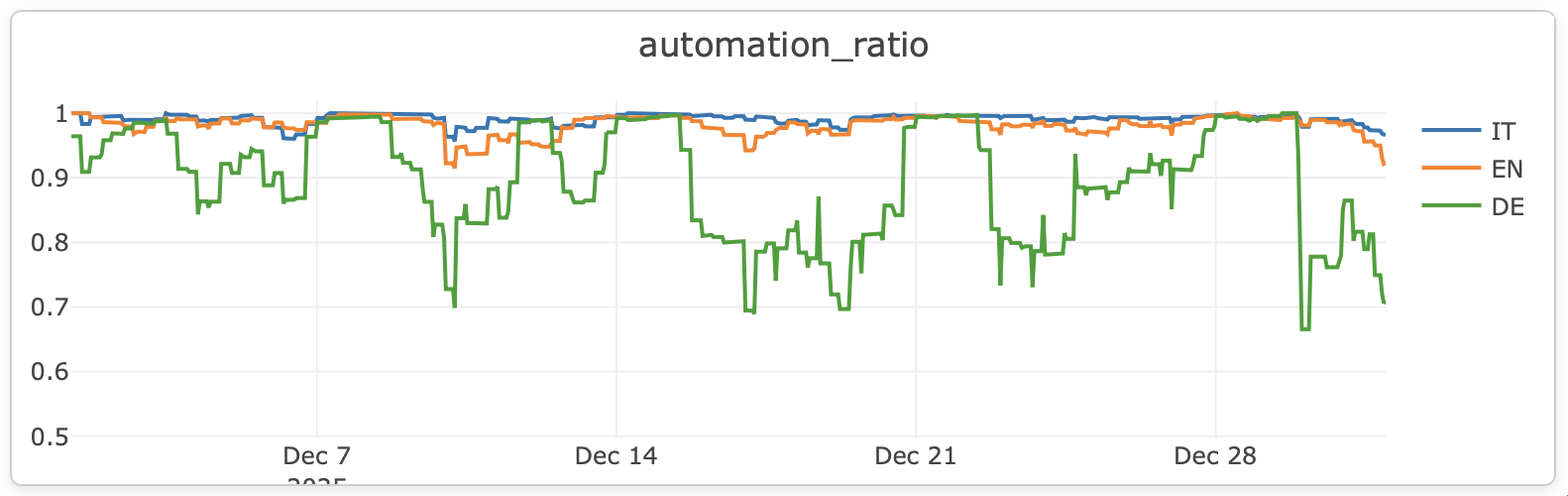}
    \caption{Screenshot of FairMon monitoring the ratio of automated moderation in the Google Play Store for Italian (IT), English (EN), and German (DE).}
    \label{fig:ratios}
\end{figure*}

As a main insight, we discovered that some languages are significantly more likely to be manually moderated, potentially indicating a gap due to insufficient numbers of fluent human moderators and an overreliance on automated moderation.
\Cref{fig:ratios} shows FairMon's output for a specification that compares the daily automation ratios of moderation decisions between different languages, i.e., how likely reported content in the given language is moderated automatically.
The figure shows this for Italian, English, and German.
Over the whole month, the likelihood of Italian and English content to be moderated automatically is consistently over 90\%, while it remains highly volatile for German, and frequently drops below 80\%.
A curious insight for German language moderation is that it approaches full automation on weekends (i.e., December 7, 14, \ldots).

Overall, the analysis with FairMon suggests that the manner and hence also the quality of content moderation may be influenced by content language and the weekday of the decision. This type of disparity has been a focus of existing legal actions made by the regulatory bodies regarding the obligations under the DSA, namely the EU Commission investigation against Meta for their actions during the 2024 Romanian Elections~\cite{meta}. The use of tools such as FairMon would enable real-time monitoring of platform-level behavior, and enable proactive enforcement and regulatory actions instead of only after-the-fact investigation.

\section{Conclusion}

FairMon extends RTLola with a probability type and dedicated operators, allowing fairness requirements to be stated concisely and naturally.
This reduces the entry barrier for users without monitoring expertise and broadens access to fairness analysis.
Paired with an interactive graphical interface that visualizes the runtime behavior, the tool supports both comprehension and practical use.
Experiments with the COMPAS case and data from the DSA Transparency database further show that the approach performs well in real-world scenarios.

\paragraph{Future Work.}
The framework is designed to be easily extensible.
Future work includes supporting a broader range of decision systems and expressing additional fairness requirements, particularly those relevant under EU regulations. We also aim to automate the extraction of API filters from specifications, reducing the need for manual configuration and further simplifying the fairness monitoring process.
Since fairness monitoring often relies on privacy-sensitive demographic and behavioral data, another promising direction is the integration of RTLola's privacy mechanisms~\cite{DBLP:conf/cav/FinkbeinerS26} into our tool to enable fairness analysis with formal privacy guarantees for monitored individuals.

\begin{credits}
\subsubsection{\ackname} This work was partially supported by the German Research Foundation (DFG) as part of TRR 248 (No.~389792660) and PreCePT (No.~521273327), and by the European Research Council (ERC) Grant HYPER (No.~101055412).

\subsubsection{\discintname}
The authors have no competing interests to declare that are relevant to the content of this article.
\end{credits}

\bibliographystyle{splncs04}
\bibliography{references.bib}

\ifthenelse{\boolean{fullversion}}{
    \clearpage
    \appendix
    \section{COMPAS Specification}\label{app:compas}
In the paper, we refer to the following (true-positive) equalized-odds specification for COMPAS with the operators presented in this paper.
The COMPAS specification without special operators can be found in the corresponding paper~\cite{full_version}.
For the line counts, we disregard empty lines, comments, and visualization annotations.

\subsection{Without Probability Operators}

\begin{lstlisting}
input event : String
input id : Int64
input group : String
input score : Int64

// Defendant Information
output days_per(i)
  spawn with id
  eval @Global(1d)@ with days_per(i).last(or:  0) + 1
  close when days_per(i) = 730
output has_re(i)
    spawn with id
    eval when id == i with event == "RECIDIVISM"
    close @Global(1d)@ when days_per(i).hold(or: 0) = 730
output tp_event(i, g, s)
  spawn with (id, group, score)
  eval @Global(1d)@
  	when days_per(i).hold(or: 0) = 730 $\wedge$ has_re(i).aggregate(over: 730d, using: $\exists$) with s > 6
  close @Global(1d)@ when days_per(i).hold(or: 0) = 730

// TP Ratio
output abs_re(g) : UInt64(*\label{ln:abs_re}*)
    spawn with group
    eval @Global(1d)@ with abs_re(g).last(or:  100) +
    tp_event.aggregate(over_instances: All(ii, ig, is => ig = g), using: count)
output abs_hr_re(g) : UInt64(*\label{ln:abs_hr_re}*)
  spawn with group
  eval @Global(1d)@ with abs_hr_re(g).last(or:  50) +
  	tp_event.aggregate(over_instances: All(ii, ig, is => ig = g), using: sum)

#[plot]
output tp_ratio(g)
    spawn with group
    eval when abs_re(g) != 0
    	with cast<UInt64, Float64>(abs_hr_re(g)) / cast<UInt64, Float64>(abs_re(g))

#[plot]
#[threshold="0.1"]
output tp_rate_diff @1d@ tp_ratio.aggregate(over_instances: all, using: max).defaults(to: 0.0) - tp_ratio.aggregate(over_instances: all, using: min).defaults(to: 0.0) > 0.1

#[plot="tp_rate_diff"]
trigger tp_rate_diff > 0.1
\end{lstlisting}

\subsection{With Probability Operators}

\begin{lstlisting}
input event : String
input id : Int64
input group : String
input score : Int64

output user_info(user)
  spawn with id when event == "SCORE"
  eval when user == id with (score, group)
  close @Local(2y)@

#[plot]
output tp_rate(r)
  spawn with group
  eval when event == "RECIDIVISM" with prob(of: score(id).hold(or: 0) > 6, given: user_info(id).exists(over: 2y) && user_info(id).1 == r)

#[plot]
#[threshold="0.1"]
output tp_rate_diff @Global(1d)@ := tp_rate.aggregate(over_instances: all, using: max).defaults(to: 0.0) - tp_rate.aggregate(over_instances: all, using: min).defaults(to: 0.0)

#[plot="tp_rate_diff"]
trigger tp_rate_diff > 0.1
\end{lstlisting}

\section{DSA Specifications}\label{app:ai_specs}

\subsection{Automation per Content Language}

\begin{lstlisting}
input event : String
input id : Int64
input group : String
input score : Int64

output user_info(user)
  spawn with id when event == "SCORE"
  eval when user == id with (score, group)
  close @Local(2y)@

#[plot]
output tp_rate(r)
  spawn with group
  eval when event == "RECIDIVISM" with prob(of: user_info(id).0.hold(or: 0) > 6, given: user_info(id).1.hold(or: "") == r)

#[plot]
#[threshold="0.1"]
output tp_rate_diff @Global(1d)@ := tp_rate.aggregate(over_instances: all, using: max).defaults(to: 0.0) - tp_rate.aggregate(over_instances: all, using: min).defaults(to: 0.0)

#[plot="tp_rate_diff"]
trigger tp_rate_diff > 0.1
\end{lstlisting}

\subsection{Decision Delay}
\begin{lstlisting}
import time
input id: Int64
input content_language: String
input automated_decision: String
input automated_detection: Bool
input platform_name: String
input application_date: String
input created_at: String

output application_date_time := time_from_rfc(application_date)
output created_at_time := time_from_rfc(created_at)

output delay_per(p)
    spawn with platform_name
    eval when platform_name == p
    with cast<UInt64,Float64>(created_at_time.time_diff(application_date_time).to_days())
    
#[plot]
output avg_delay_per(p)
    spawn with platform_name
    eval @1h@ with delay_per(p).aggregate(over: 1d, using: avg).defaults(to: 0.0)
\end{lstlisting}

\subsection{Automation Ratios}

\begin{lstlisting}
input id: Int64
input content_language: String
input automated_decision: String
input automated_detection: Bool
input platform_name: String

output automation (p, dec, det)
    spawn with (platform_name, automated_decision, automated_detection)
    eval with prob(of: automated_decision == dec, given: automated_detection == det)

output automation_plot(p, dec, det)
    spawn with (platform_name, automated_decision, automated_detection)
    eval @60s@ with automation(p, dec, det).hold(or: 0.0)

#[plot]
output tiktok(dec, det)
    spawn with (automated_decision, automated_detection)
    eval @60s@ with automation_plot("TikTok", dec, det).hold(or: 0.0)

#[plot]
output shopify(dec, det)
    spawn with (automated_decision, automated_detection)
    eval @60s@ with automation_plot("Shopify", dec, det).hold(or: 0.0)

#[plot]
output idealo(dec, det)
    spawn with (automated_decision, automated_detection)
    eval @60s@ with automation_plot("Idealo", dec, det).hold(or: 0.0)

#[plot]
output ebay(dec, det)
    spawn with (automated_decision, automated_detection)
    eval @60s@ with automation_plot("eBay", dec, det).hold(or: 0.0)
\end{lstlisting}

\section{Query Delay Details}

We retrieve the data for 26.11.2025 using a page size of 1000.
If a quota limit is reached, the process pauses and resumes with one-second intervals until the quota becomes available again.

\begin{center}
\begin{tabular}{lrrrr}
\hline
Scenario & Pages & Full time (s) & Hits & Mean latency (s) \\
\hline
1d\_all & 5501 & 6662.741 & 5500000 & 0.928 \\
1d\_instagram      & 379  & 484.664  & 378449  & 0.986 \\
\hline
\end{tabular}
\end{center}

}{}

\end{document}